# Visualising Information Flow in Large Language Models via Diffusion Tensor Imaging

Thomas Fabian

Technical University of Darmstadt

thomas.fabian@tu-darmstadt.de

## Abstract

Understanding how large language models (LLMs) represent natural language is a central challenge in natural language processing (NLP) research. Many existing methods extract word embeddings from an LLM, visualise the embedding space via point-plots, and compare the relative positions of certain words. However, this approach only considers single words and not whole natural language expressions, thus disregards the context in which a word is used. Here we present a novel tool for analysing and visualising information flow in natural language expressions by applying diffusion tensor imaging (DTI) to word embeddings. We find that DTI reveals how embedding space representations change between tokens. Tracking these changes within the layers of an LLM allows for comparing different model structures and could potentially reveal opportunities for pruning an LLM's under-utilised layers. Our results show that our visualisation method permits novel insights into how LLMs represent actual natural language expressions, extending the comparison of isolated word embeddings and improving the interpretability of NLP models.

# Introduction

Modern natural language processing (NLP) models, or large language models (LLMs), depend on representations of words in high-dimensional embedding spaces. LLMs generate a high-dimensional word embedding with rich linguistic information for every word in natural language. However, interpreting the information captured in these embeddings is challenging because individual dimensions lack intuitive meaning (Bender & Koller, 2020), and the process of learning these representations from large amounts of textual data remains opaque (Lipton, 2018; Rudin, 2019). Thus, enhancing the interpretability of NLP models has become a scientific endeavour of its own, whose goals are understanding how LLMs represent linguistic information and decoding this information. To better understand and control artificial intelligence, especially for building trust in these models and ensuring their responsible deployment, it is essential to understand how information is represented in LLMs. To achieve this, there are numerous approaches for visualising high-dimensional embeddings of LLMs.

A common approach to visualising word embeddings are projections or scatter-plots (Ahlberg & Shneiderman, 1994; Maaten & Hinton, 2008; McInnes et al., 2020; Smilkov et al., 2016; Weaver, 2004). These can be used to compare the locations of words in the embedding space, or to find the nearest neighbours or semantic clusters. The relational connections between word embeddings can provide valuable insights into how LLMs represent information, such as the fact that subtracting the embedding for *man* from the embedding for *woman* reveals a direction vector encoding femininity in the embedding space (Mikolov et al., 2013). Similarly, semantic clusters can be identified by preserving local neighbourhoods through algorithms such as t-SNE (Maaten & Hinton, 2008), while manifold methods such as UMAP (McInnes et al., 2020) retain more structural details in the low-dimensional layout. However, scatter plots exhibit two major shortcomings. First, in the plots they create, the axes do not have any intuitive meaning due to the dimensionality reduction from the high-dimensional embedding space to only a few dimensions suitable for visualisation. Thus, statements about embedding spaces rest on linear projections or scatter plots that can only be interpreted by taking relative positions into account. Second, existing visualisation methods can only compare the embeddings of isolated words, without providing insights into how information representations vary in the context of actual expressions of syntactically and semantically connected words. Hence, they cannot visualise how embeddings change between words in a sentence as it occurs in natural language.

Newer methods address these shortcomings by providing interpretable scales and visualising how the same word can behave in different contexts and how the word's meaning can vary (Berger, 2020; Engler et al., 2023; Escolano et al., 2019; Sevastjanova et al., 2022, 2025; Yan et al., 2025). For example, SensePOLAR (Engler et al., 2023) transforms the contextual word embeddings of an LLM into interpretable semantic scales that describe the meaning of a token in a given context. Thus, this method provides scale interpretability and context-awareness. However, like other existing methods in this line of research, SensePOLAR only focuses on the meaning of single tokens, does not visualise all the layers of an LLM, and cannot capture how the embeddings of an LLM change across tokens in natural language expressions.

In the field of neuroscience, visualising changes, i.e., flows of liquids, is well established. Diffusion tensor imaging (DTI) has been used for decades to visualise diffusivity in the human brain (Basser et al., 1994; LeBihan et al., 2001; Westin et al., 2002). The idea behind DTI is to calculate a $3 \times 3$ diffusion tensor around each voxel, which is then visualised using a diffusion ellipsoid. The lengths of the ellipsoid axes show the strength and direction of water diffusivity in tissue. In this context, isotropic diffusion refers to diffusivity being equally strong in all directions, while anisotropic diffusion denotes the diffusion being stronger along one axis than along the other axes. Since water diffusivity, or flow, is highest along the primary fibre direction, the direction of the anisotropic diffusion ellipsoids indicates the alignment of fibres in the brain. More notably, DTI not only allows for investigating diffusivity in the brains of healthy individuals, but is also particularly useful for investigating which alterations of diffusivity occur when people suffer from brain disorders (Arbabshirani et al., 2017; Pini et al., 2016; Wood et al., 2024).

As with the human brain, understanding the activities of individual structures in LLMs can help identify the integral components and underutilised parts of the system. In NLP research, the goal of structured pruning, i.e., the removal of redundant parameters or superfluous components of a neural network, is to reduce model complexity with as little performance loss as possible (Hassibi et al., 1993; LeCun et al., 1989). Effective pruning of NLP models is of considerable practical importance, as operating a model after training, known as inference, accounts for approximately 60-70% of an LLM's computing and energy requirements (Desislavov et al., 2023; Mavromatis et al., 2024; Patterson et al., 2022; Wu et al., 2022). Numerous studies introduce methods for reducing the model complexity of LLMs, among other things, to reduce the computational effort and energy requirements during inference (Blalock et al., 2020; Fan et al., 2019; Tao et al., 2023; Teerapittayanon et al., 2016; Varshney et al., 2024; Xia et al., 2022). In this context, metrics of which layers in an LLM observe little change during language processing could be used to reveal the need for pruning or to optimise pruning methods by indicating remaining inefficiencies or explaining performance losses.

Based on DTI and its capability of highlighting alterations of flow, we propose a novel visualisation method to depict directional relations between word embeddings, indicating the information flow in natural language expressions of any length. For this methodological transfer from neuroscience to LLMs, we define information flow as the changes between the embedding values of an LLM during language processing, drawing on definitions of previous research in this field (Bogdan, 2025; Chang et al., 2026; DeRose et al., 2020; Sevastjanova et al., 2022; Skean et al., 2025; Sreehari et al., 2025; X. Zhang et al., 2024). According to our definition, information flow occurs when the value of an embedding is not identical to its neighbourhood. Inversely, if adjacent embeddings are identical, we consider there to be no information flow as the information in both embeddings is the same and no change can be observed between them.

To evaluate information flow via diffusion ellipsoids, we propose a tool for **D**iffusion-**O**riented **N**atural-language **A**nisotropy and **L**ayer-wise **D**irectionality **D**iagnostics (DONALD-D). We visualise the flow of information through the words of a sentence using diffusion ellipsoids, which we calculate based on word embeddings. DONALD-D rests on the idea of concatenating the embeddings of words in a sentence to a two-dimensional matrix, applying DTI to calculate structure tensors within this matrix, and visualising the information flow across words with diffusion ellipsoids. With this procedure, DONALD-D transforms abstract embedding patterns into human-interpretable images of information flow within natural language expressions of any length.

The contribution of this work is the introduction of a novel visualisation method for LLMs' embedding spaces based on diffusion ellipsoids. DONALD-D leverages the interpretability of DTI to visualise the information flow between words across all layers of an LLM in the context of natural language expressions, relationships that cannot be simultaneously visualised using existing methods. DONALD-D improves the interpretability of LLMs by providing four features that existing visualisations do not combine: (1) the axes of the visualisation can be interpreted quite intuitively, (2) the relationships between embeddings can be visualised in the actual context of real natural language expressions, (3) all tokens are visualised with equal importance, i.e., the visualisation does not only explain the meaning of a single token in a specific context, and (4) all layers of an LLM are visualised. Due to its direct applicability to natural language expressions of any length, DONALD-D allows for new investigations into the structure of LLM embedding spaces. By enabling the assessment of layer utilisation in embedding spaces, DONALD-D allows the identification of opportunities to prune individual under-utilised layers within an LLM or explain why a pruned model loses performance.

# Methods

## Model structure

**Input processing.** DONALD-D operates with any common hidden-state format used in LLMs. The numbers of layers $L$, tokens $T$, and hidden units $H$ usually form an embedding space as stacked layer arrays or per-token stacks with shapes $(\mathrm{L}, \mathrm{T}, \mathrm{H})$ and $(\mathrm{T}, \mathrm{L}, \mathrm{H})$, respectively. This applies to all models available on Hugging Face (Wolf et al., 2020). For unconventionally shaped embedding spaces, DONALD-D includes an automated layout detection by axis sizes. Although DONALD-D can process all embeddings, we focus on the transformer-block outputs and exclude the embedding output as it encodes conceptually different information (Ethayarajh, 2019; Tenney et al., 2019). Note that for one sentence, the number of tokens $T$ can vary between models as the tokenisation processes differ. The first transformation is collapsing the hidden-unit axis $H$ by an arithmetic mean for layer $\mathrm{i} \in \{0, \dots, \mathrm{L}-1\}$ and token $\mathrm{j} \in \{0, \dots, \mathrm{T}-1\}$:

$$\mathrm{M}_{\mathrm{i,j}} = \frac{1}{\mathrm{H}} \sum_{\mathrm{h}=1}^{\mathrm{H}} \mathrm{E}_{\mathrm{i,j,h}}.$$

This yields a matrix $\mathrm{M} \in \mathbb{R}^{\mathbb{L} \times \mathbb{T}}$ containing one value for each layer-token combination. Despite losing information, we opt for the arithmetic mean as the individual values of hidden units are not intuitively interpretable. Visualising the hidden units would be possible with DTI. However, including the hidden units would not improve the interpretability of the visualisations as this third dimension impair readability due to overloading and introducing the general ambiguities in three-dimensional visualisations (Cleveland & McGill, 1984; Tufte, 1983). Furthermore, we decide against principal component analysis for dimension reduction, as the results would be rendered intuitively uninterpretable.

After collapsing the hidden units, we apply row-wise min-max normalisation to better compare information flow within layers and between tokens. By applying min–max normalisation per each row $i$

$$\mathrm{M}_{\mathrm{i},\cdot} = \frac{\mathrm{M}_{\mathrm{i},\cdot} - \min_{\mathrm{j}} \mathrm{M}_{\mathrm{i},\mathrm{j}}}{\max_{\mathrm{j}} \mathrm{M}_{\mathrm{i},\mathrm{j}} - \min_{\mathrm{j}} \mathrm{M}_{\mathrm{i},\mathrm{j}}}$$

the values of each row are normalised to the interval $[0,1]$. By equalising the range of each layer, the resulting tensors become comparable across layers even if one layer has systematically larger values than another. In combination with the preserved token-wise variation, the calculated anisotropy is comparable between layers, yet sensitive to variations between tokens, serving the goal of visualising information flow in natural language.

For completeness and possible other applications, DONALD-D also provides options for column-wise and global min-max normalisation prior to gradient estimation, but both impair the visualisation and detectability of information flow. Column-wise min-max normalisation removes cross-token variations, i.e., information flow, altogether. Global normalisation minimises contrast as outliers dominate the normalisation interval, compressing the remaining matrix entries into a narrow range of values. In the subsequent gradient estimation, these similar values result in small gradients and near-isotropic tensors across the entire matrix, rendering the detection of information flow impossible.

**Structure tensor estimation.** To estimate the structure tensors, we first determine the gradients in matrix $M$. DONALD-D calculates the gradients $\partial_x M$ (horizontal/token-to-token) and $\partial_y M$ (vertical/layer-to-layer) as discrete spatial derivatives on the basis of centred finite differences. The gradients indicate the slope of the spatial variation in the values for each token-layer combination. As the previous normalisations yield unit grid spacing for matrix $M$, we can compute its derivative fields element-wise as

$$\partial_x M_{i,j} = \frac{M_{i,j+1} - M_{i,j-1}}{2},$$

$$\partial_y M_{i,j} = \frac{M_{i+1,j} - M_{i-1,j}}{2}.$$

Next, we apply Gaussian smoothing to the derivative fields to reduce noise sensitivity in the separately calculated derivatives. The smoothing is implemented as a convolution of each derivative field, $\partial_x M$ and $\partial_x M$, with an isotropic Gaussian kernel. For applications that require stronger smoothing along either layers or tokens, an anisotropic smoothing kernel can be applied at this point. The derivative fields for the horizontal and vertical gradients are used to calculate the raw second-moment components

$$J_{xx} = (\partial_x M)^2,$$

$$J_{xy} = (\partial_x M)(\partial_y M),$$

$$J_{yx} = (\partial_y M)(\partial_x M),$$

$$J_{yy} = (\partial_y M)^2,$$

which indicate directional change, or flow, across the horizontal, vertical, and diagonal axes. We apply Gaussian smoothing to stabilise the noise-sensitive raw second-moment components and obtain the coherent local directionality of neighbourhoods.

Lastly, the structure tensors are assembled from the smoothed second-moment components. Each cell of the initial token-layer matrix $M$ can be represented as a combination of the directional flows through it. For each cell, the information flow is captured in a symmetric $2 \times 2$ structure tensor

$$J_{i,j} = \begin{pmatrix} \mathrm{J_{xx}(i,j)} & \mathrm{J_{xy}(i,j)} \\ \mathrm{J_{yx}(i,j)} & \mathrm{J_{yy}(i,j)} \end{pmatrix} + \varepsilon I,$$

with $\varepsilon = 10^{-12}$ added to ensure numerical stability. With this procedure, DONALD-D performs a numerically stable estimation of structure tensors that capture the local pattern of information flow for each token-layer combination of the LLM's embedding space. Although the pattern of information flow is calculated locally, it depends on the whole input as transformer models are based on the idea of every input paying attention to all other inputs (Vaswani et al., 2017). Thus, if words are added to the end of a sentence, the first tokens' representations in the embedding space will change to capture the new context. One could calculate the structure tensors for longer sequences but we suspect the added insight to be limited as long-range dependencies between tokens are already part of a token's embedding space representation.

With the tensor components $\mathrm{J_{xx}(i,j)}$ and $\mathrm{J_{yy}(i,j)}$, we also determine the utilisation rate of every layer, i.e., the fraction of information diffusion along the token-to-token direction, as

$$\mathrm{U_i} = \frac{1}{\mathrm{N}} \sum_{\mathrm{j=1}}^{\mathrm{N}} \frac{\mathrm{J_{xx}(i,j)}}{\mathrm{J_{xx}(i,j) + J_{yy}(i,j)} + \varepsilon},$$

with $\varepsilon = 10^{-12}$ for avoiding division by zero. Note that including the diagonal tensor components $\mathrm{J_{xy}(i,j)}$ and $\mathrm{J_{yy}(i,j)}$ is not necessary as these components themselves are calculated from the vertical and horizontal gradients. For every layer, the utilisation rate $U$ denotes the average fraction of diffusion that takes place between tokens and not layers. A low utilisation rate indicates little change across tokens within a layer. Thus, it suggests that the respective layer might be redundant in the embedding space as the information contained in that layer does not change between tokens. In case most differences between tokens are captured by other layers, the under-utilised layer could be removed since the ratio of added informational value to required computational demand is low compared to the rest of the LLM's layers.

**Anisotropy calculation.** Calculating the anisotropy of each cell in matrix $M$ necessitates the eigen-decomposition of each local structure tensor J as

$$J_{i,j} v_k(i,j) = \lambda_k(i,j)\, v_k(i,j),$$

with $k = 1, 2$. Diagonalising structure tensor $J_{i,j}$ yields the eigenvectors $v_1(i,j)$ and $v_2(i,j)$, from which the principal eigenvector $v_1(i,j)$ indicates the dominant direction of local flow, and the eigenvalues $\lambda_1(i,j) \geq \lambda_2(i,j)$, which quantify the amount of energy aligned with each direction. From the eigenvalues, we calculate the degree of anisotropy following established practice (Bigun & Granlund, 1987; Weickert, 1999) as

$$A(i,j) = \frac{\lambda_1(i,j) - \lambda_2(i,j)}{\lambda_1(i,j) + \lambda_2(i,j) + \varepsilon},$$

with $\varepsilon = 10^{-12}$ avoiding division by zero. This normalised difference expresses the anisotropy in the interval $[0,1)$, with $A = 0$ indicating no preferred direction, i.e., isotropy (Fig. 1, left). Higher values suggest more anisotropic diffusion, with the information flow being more pronounced along one direction than the other (Fig. 1, right).

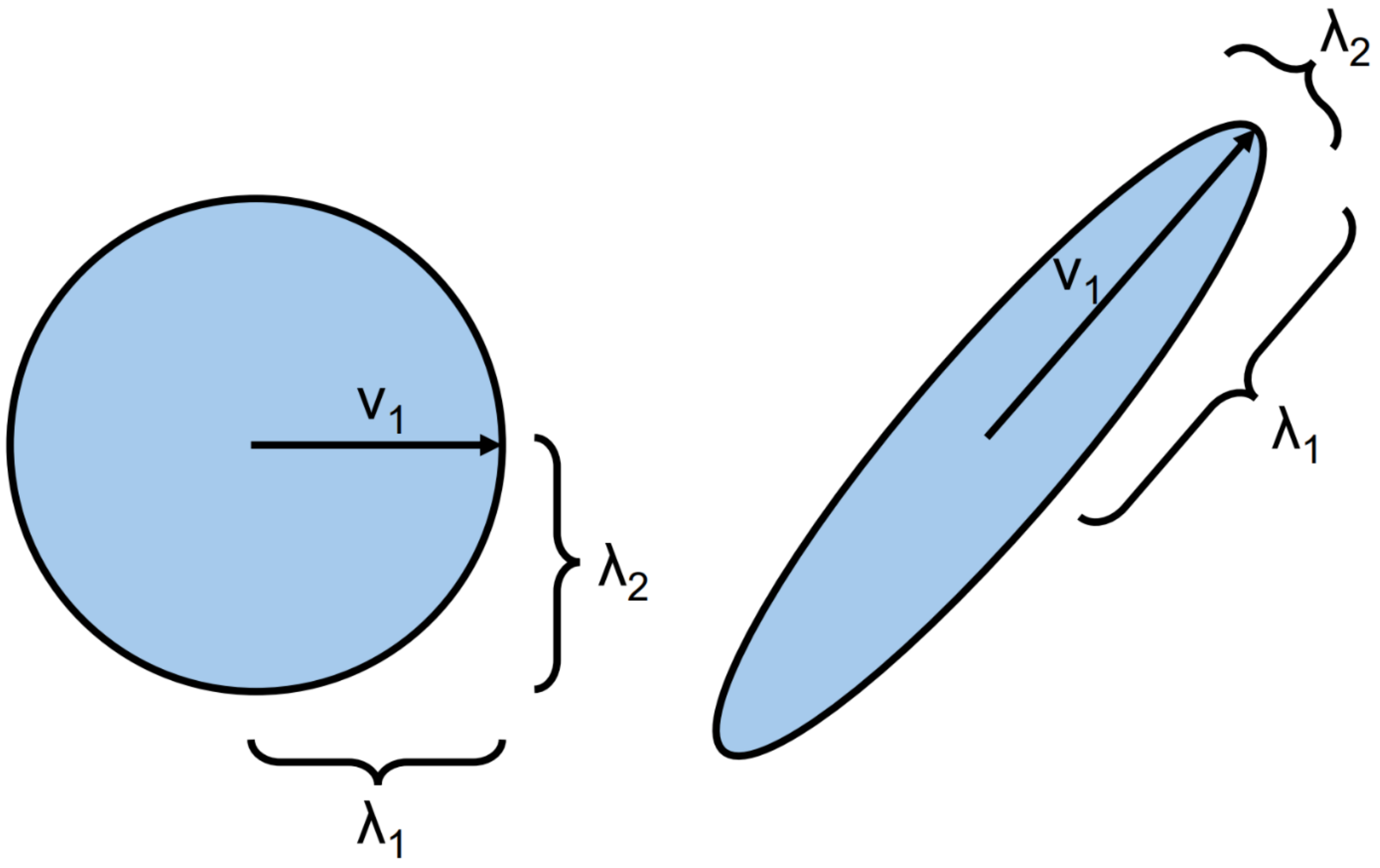

**Figure 1. Isotropic and anisotropic diffusion.** For isotropic diffusion (left), the eigenvalues $\lambda_1$ and $\lambda_2$ are equal. The degree of anisotropy (right) increases with the difference between the eigenvalues as a larger fraction of directional strength runs along the principal eigenvector $v_1$.

By determining the eigenvectors and anisotropy, we obtain a principal vector field $v_1(x, y): \{0, \dots, L-1\} \times \{0, \dots, T-1\} \rightarrow \mathbb{R}^2$ and the anisotropy scalar field $A(x, y): \{0, \dots, L-1\} \times \{0, \dots, T-1\} \rightarrow [0,1)$. Thus, for each combination of token and layer, DONALD-D provides the dominant direction as a two-dimensional vector and the fraction of energy that is applied to that direction.

This anisotropy measure is the two-dimensional analogue of fractional anisotropy (Basser et al., 1994), a three-dimensional anisotropy measure commonly used in DTI for neuroimaging. Alternative to our measure, one could determine anisotropy with different methods like tensor-shape decompositions (Westin et al., 2002) to capture the shape of diffusion with more detail. We opt for normalised differences in DONALD-D to get a simple and intuitively interpretable visualisation of the structure tensors.

As the principal vector field and anisotropy scalar are calculated for each cell, they are only defined for the grid centres and can thus only be queried with integers. We included an interpolation so, technically, DONALD-D can be queried for principal flow direction and anisotropy at any continuous point within the matrix boundaries. However, as values off the grid centres lack interpretability, we focus on the discrete matrix structure to refer to distinct token-layer combinations.

**Visualisation.** We visualise the structure tensors as diffusion ellipsoids at the centre of every cell in the token-layer matrix. The ellipsoid is oriented in the direction of the principal eigenvector $v_1$ and its semi-axes are proportional to $\sqrt{\lambda_1}$ and $\sqrt{\lambda_2}$. Thus, each ellipsoid encodes the dominant direction and the directional strength, with a radially symmetric ellipsoid indicating isotropic diffusion. We apply a global scale factor to ensure that all ellipsoids fit into the corresponding matrix tile in the visualisation.

To visually highlight information flow directions, we colour the tiles of each token-layer combination according to their dominant direction of flow. The colourmap we apply is $\pi$-periodic, i.e., an angle $\theta$ is assigned the same colour as $\theta + 180°$, as the calculated structure tensors do not indicate the direction of change. This property of DTI should pose no problem to the visualisation of information flow in natural language since all tokens have a clearly defined sequence, leaving no possibility for changes in the opposite direction of this sequence.

Given that all information must flow in the token sequence direction, we colour both diagonal axes identically as they equally reflect a deviation from the token-to-token information flow. Information flow running parallel to the token sequence is coloured red, diagonal deviations yellow, and flows perpendicular to the token sequence blue. To encode the directional strength in the colouring, the transparency of each tile is directly proportional to its degree of anisotropy.

Thus, token-layer combinations with low directional strength, i.e., near-isotropic information diffusion, are visualised as almost white tiles.

The final visualisations created by DONALD-D thus comprise a colour-coded image of rectangular tiles, overlayed with ellipsoids encoding the directional strength of the local information diffusion. The colouring allows the plot to be read like a heatmap with red tiles indicating strong token-to-token changes within a layer and blue tiles indicating no information flow between tokens. The model's parameters are set to balance the visualisation's readability and the numerical accuracy. For other visualisation purposes, the parameters can be adjusted for optimal results. Concerning DONALD-D's complexity, visualising how information flows through an LLM's word embeddings scales linearly with the length of natural language expressions. Each processing step operates with independent $2 \times 2$ matrices that are calculated from matrix $\mathrm{M} \in \mathbb{R}^{L \times T}$ As the number of layers $L$ is constant for any given LLM, the complexity for analysing a natural language expression containing $T$ tokens is $O(T)$. We created the python package donald-d to allow any researcher to easily and directly run the model. The code for DONALD-D is publicly available on GitHub (https://github.com/Thomas-1103/donald-d).

## Model validation

**Corpus preparation.** We perform a corpus analysis to validate whether DONALD-D is suitable for identifying under-utilised transformer layers and to evaluate whether the identified layers can be removed with limited loss in language modelling quality. All experiments use the Brown corpus (Francis, 1965; Kučera & Francis, 1967), an established and self-contained benchmark corpus that is publicly available and diverse in style and genre. Furthermore, the Brown corpus can be split deterministically and its use does not rely on proprietary preprocessing.

The Brown corpus is split into three disjoint partitions to obtain clearly separated corpus data for (1) scoring the utilisation rates of layers to decide which layers to prune, (2) fitting the pruned model, and (3) evaluating performance. We opt for a 15% scoring, 70% training, and 15% testing split. To prevent overlap between tasks, Brown file identifiers are randomly partitioned into three disjoint subsets. We split the data according to files rather than sentences since sentences from the same text often share topic and stylistic context. Thus, splitting at sentence level would introduce leakage and inflate the generalisability of results (Elangovan et al., 2021; Joeres et al., 2025; Zhao et al., 2026). The split is conducted with a fixed random seed to ensure reproducibility.

**Model pruning.** The scoring split of Brown is used to assess the utilisation of the LLM's layers. For every text in the scoring split, the model calculates the utilisation rate for every layer. The utilisation rates are then aggregated across the split to obtain the mean, standard deviation, and 95% confidence interval for each layer. The layers' mean utilisation rates for the scoring split are used to identify which layers should be pruned. For this decision, a threshold of minimal utilisation must be chosen. We run the analyses with various possible thresholds from 5% to 50% to evaluate the threshold's influence on the pruning process and to assess what utilisation rate indicates a good cost-benefit profile for keeping a layer. Layers with mean utilisation rates below the respective threshold are pruned from the model.

The pruned model is constructed by instantiating the original LLM with a reduced number of transformer blocks, i.e., the transformer blocks with above-threshold utilisation rates. Furthermore, the token embeddings of the original model are copied, as well as the positional embeddings and final layer norm. This procedure constitutes structural pruning and not merely masking as layers are removed altogether and the resulting model has fewer parameters and lower computational depth. Alternative dynamic gating or routing (Krishnamurthy et al., 2022; Shazeer et al., 2017; Wang et al., 2018; Zhou et al., 2022) would have preserved the original architecture but would not have yielded the same reduction in model size and inference cost. We implemented a practical safeguard for the pruning step: the implementation always retains the single most utilised layer so that the pruned model is still valid even if the threshold were to suggest the removal of all layers.

**Model readjustment.** After pruning, we perform teacher-student readjustment on the Brown training split with the pruned model as the student and the full model as the teacher. This distillation is necessary as pruning, specifically the removal of whole layers, can lead to the loss of specific functionalities (Gao et al., 2026; Liu et al., 2025; Pochinkov & Schoots, 2024; Shrestha et al., 2026; H. Zhang et al., 2024). To assess which information is still present in the pruned model, we readjust the model to the compressed structure. During readjustment, the loss that is optimised for combines a causal language modelling loss and a distillation loss over logits:

$$\mathcal{L} = \alpha\mathcal{L}_{CE} + (1-\alpha)\mathcal{L}_{KL}.$$

The cross-entropy loss $\mathcal{L}_{CE}$ is computed from the next-token targets in the input sequence. The distillation loss $\mathcal{L}_{KL}$ captures the temperature-scaled Kullback-Leibler (KL) divergence (Kullback & Leibler, 1951) between the teacher and student output distributions. A temperature of 2.0 is used as a conservative distillation choice for softening the logits without rendering the distribution uninformative. The weight $\alpha = 0.5$ balances the influence of direct language-modelling fit and teacher guidance.

When readjusting the pruned model, we use a learning rate of $5 \times 10^{-5}$ for light adaptation of a pretrained language models. This learning rate allows for updating the student while being conservative enough not to destroy its inherited structure. The number of distillation epochs is set to 1 as we aim for slightly readjusting the model. More distillation epochs could lead to overfitting on the training data and would also be more akin to retraining the pruned model from scratch. A single epoch appears sufficient to test how much information the pruned architecture retains. The distillation parameters were selected as conservative defaults rather than empirically determined optima.

**Evaluation.** The pruned model is evaluated on the Brown test split, i.e., on data completely disjoint from the scoring and training splits. To evaluate the benefits of our pruning procedure, we calculate the parameter count and approximate the inference time. The parameter count provides a direct measure of the extent of compression achieved by the pruning, while throughput gives a practical indication of computational cost. We use these two metrics to assess the reduction in size and complexity achieved by the utilisation-based pruning. However, pruning also entails a drawback: as the model is smaller, it tends to contain less information and, thus, have diminished performance compared to the unpruned model. For performance evaluation, we report the negative log-likelihood. It is computed with a sliding-window over the test corpus since the model's context length is finite, making it necessary to chunk long texts into segments.

# Results

## Qualitative results

For demonstrating how DONALD-D reveals model-dependent differences in information flow, we visualise the information flow in BERT (Devlin et al., 2019), Longformer (Beltagy et al., 2020), GPT-2 (Radford et al., 2019), and PEGASUS (J. Zhang et al., 2020) for an exemplary natural language expression. We choose the first sentence of the first file in the Brown corpus (Francis, 1965; Kučera & Francis, 1967) as the natural language expression. The four selected LLMs represent a classical encoder-only model (BERT), an encoder-only model with sparse attention to capture long sequences (Longformer), a decoder-only model (GPT-2), and an encoder-decoder hybrid (PEGASUS). For this work, we use only license-free models available on Hugging Face (Wolf et al., 2020) to ensure unrestricted reproducibility. However, DONALD-D can be applied to any LLM and any type of embedding as they all encode words as mathematical abstractions, whose concatenation can be examined for information flows via DTI. Figure 2 shows DONALD-D's visualisation of information flows for the same exemplary input in four different LLMs. All results, including the full list of quantitatively determined utilisation rates per layer for each model, and the code needed to create them can be found in the study's Open Science Framework repository (https://osf.io/rc3ts).

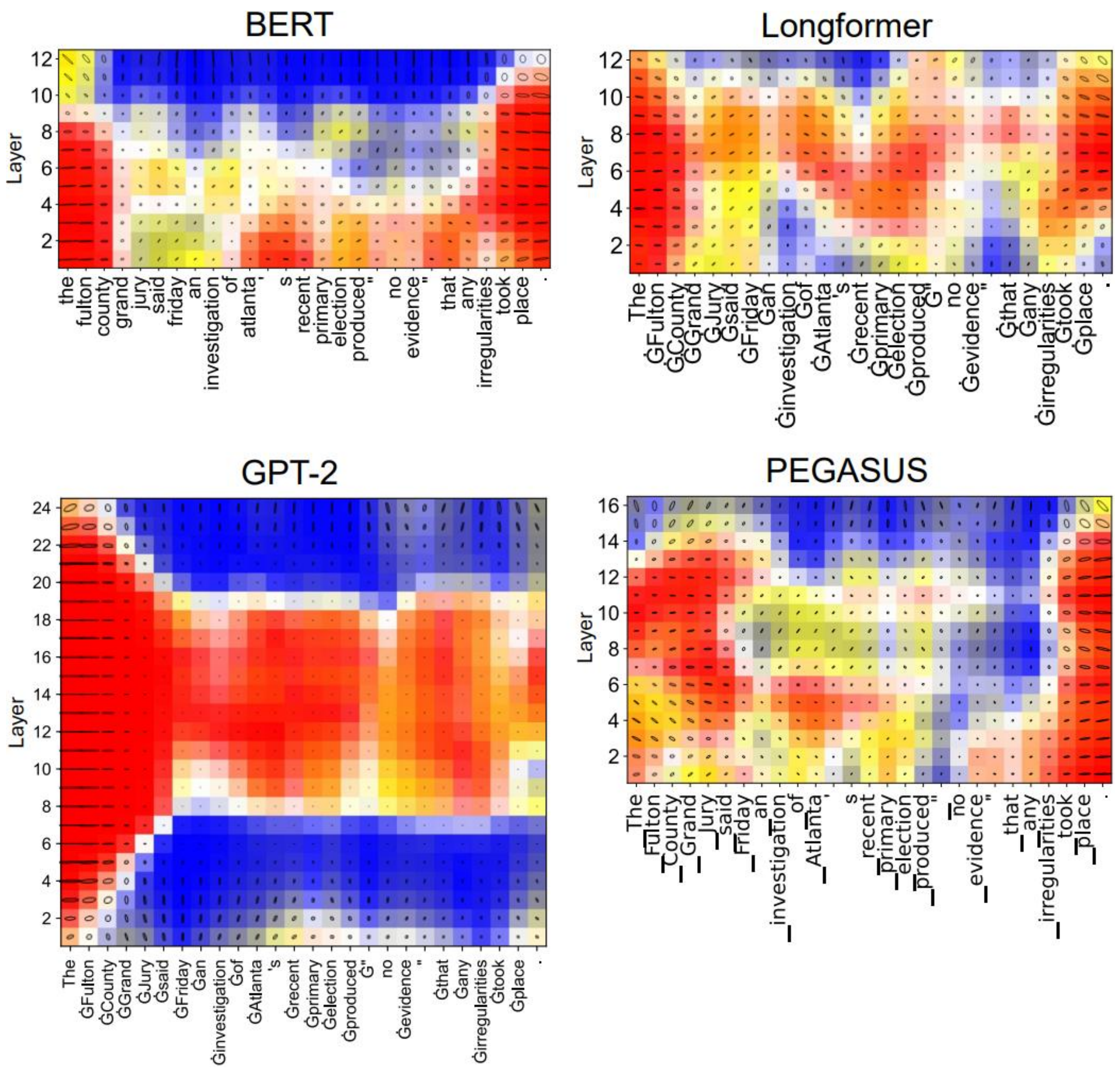

**Figure 2. Information flow of an exemplary natural language expression in four different LLMs.** The same natural language expression is provided to BERT (Devlin et al., 2019), Longformer (Beltagy et al., 2020), GPT-2 (Radford et al., 2019), and PEGASUS (J. Zhang et al., 2020). Each model tokenises and provides word embeddings for the sentence: *The Fulton County Grand Jury said Friday an investigation of Atlanta's recent primary election produced "no evidence" that any irregularities took place.* The x-axis shows the tokens, the number of which can vary between models due to different tokenisation processes. The y-axis shows the layers of each model. The ellipsoids within each tile encode the direction of information diffusion. Red tiles indicate token-to-token information flow. Blue tiles indicate no information flow along the token sequence. No preprocessing was performed to analyse natural language expressions and create visualisations with DONALD-D.

The visualisations show how the two encoder-only models BERT and Longformer differ in their information flow (Fig. 2, top). While BERT utilises layers one to eight the most, the last four layers only slightly change across most tokens. In Longformer, the same example sentence yields more evenly distributed information flows, with each section of the layers being utilised for capturing changes across tokens in different parts of the sentence. The utilisation rates deliver similar results as they more evenly distributed across layers in Longformer (mean = 60.91 %, SD = 16.38 %) than BERT (mean = 48.42 %, SD = 24.40 %). The encoder-decoder model PEGASUS (Fig. 2, bottom right) shows information flows similar to those of BERT (Fig. 2, top left), with the highest layers being under-utilised. The average utilisation rate per layer and the deviation by layer in PEGASUS also appear similar to the encoder-only models (mean = 51.43 %, SD = 16.78 %).

For the decoder-only model GPT-2 (Fig. 2, bottom left), all layers display strong information diffusion along the first five tokens. However, from the sixth token onwards, the layers' utilisation shows pronounced differences. Layers one to seven and twenty to twenty-four show little information flow, as more than two-thirds of the information diffuses perpendicular to the token sequence. Layers eight to nineteen exhibit the majority of changes between tokens, indicating that it is mainly these layers that allow the model to distinguish one token's representation from the surrounding tokens. For GPT-2, the average layer utilisation is comparable to the other models but the utilisation variation by layer is considerably higher than for the encoder-only models (mean = 51.85 %, SD = 33.70 %).

As DONALD-D is capable of visualising natural language expressions of any length, we also perform exemplary visualisations of expressions with different lengths to demonstrate this feature. We use the first three sentences in the Brown corpus (Francis, 1965; Kučera & Francis, 1967) for this illustrative purpose. The visualisations are created for GPT-2, as the model comparison revealed the most pronounced layer-wise differences in information flows. The visualisations are shown in Figure 3.

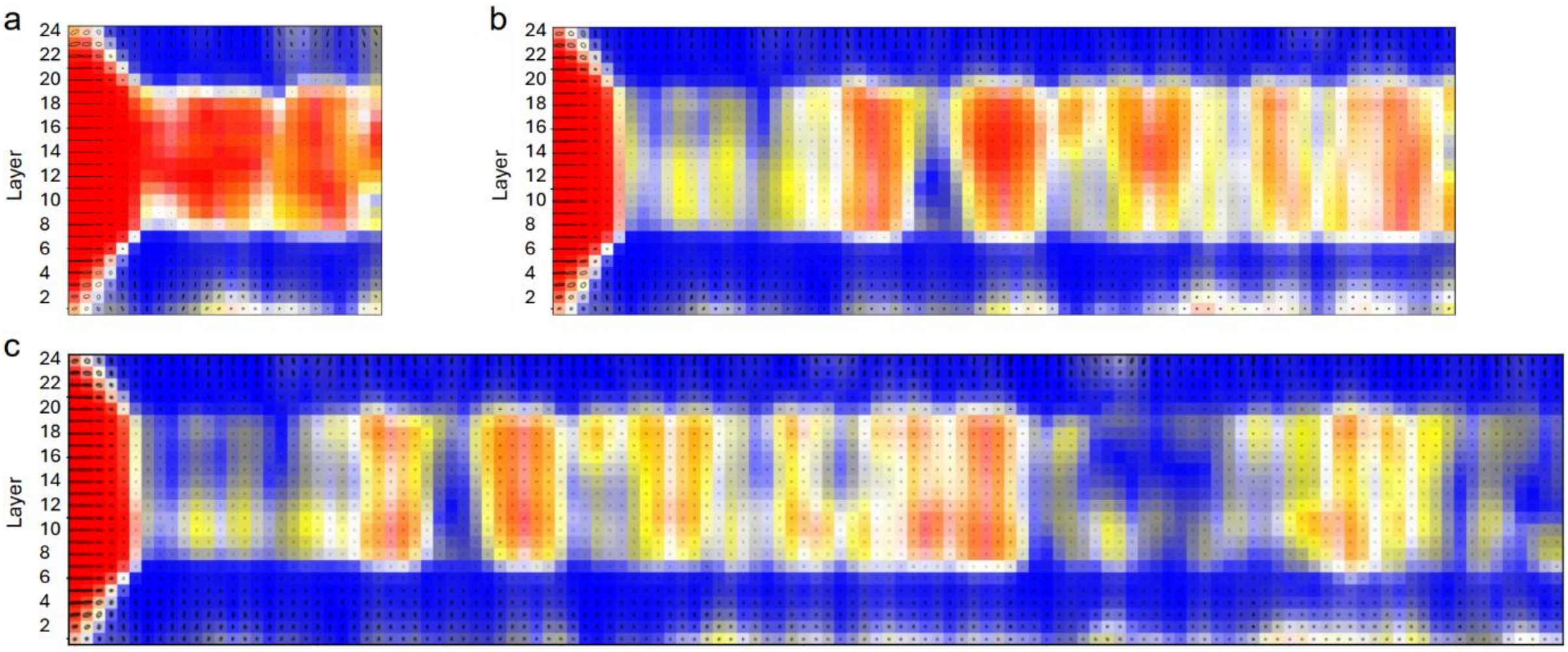

**Figure 3. Information flow of a long natural language expression in GPT-2.** Information flows in GPT-2 (Radford et al., 2019) when processing long natural language expressions, illustrated by the first three sentences in the Brown corpus (Francis, 1965; Kučera & Francis, 1967). Each column represents a token and each row one of GPT-2's 24 layers. **a)** Shows the visualisation of the first sentence with 26 tokens. **b)** Shows the information flows within the first two sentences with a total of 75 tokens. **c)** Shows the visualisation of information flows in the first three sentences with a total of 123 tokens. The first three sentences in the Brown corpus are: *The Fulton County Grand Jury said Friday an investigation of Atlanta's recent primary election produced "no evidence" that any irregularities took place. The jury further said in term-end presentments that the City Executive Committee, which had over-all charge of the election, "deserves the praise and thanks of the City of Atlanta" for the manner in which the election was conducted. The September-October term jury had been charged by Fulton Superior Court Judge Durwood Pye to investigate reports of possible "irregularities" in the hard-fought primary which was won by Mayor-nominate Ivan Allen Jr.*.

The visualisations show that the utilisation of the LLM remains qualitatively similar in long expressions, with the upper and lower layers exhibiting little token-to-token differences while the central layers show high information flow. It is also noticeable, however, that the flow of information in the centre is stronger for a single sentence (Fig. 3a) than for two (Fig. 3b), and continues to decrease as the length of the expression increases (Fig. 3c). This is to be expected, as the embeddings of LLMs take the entire processed sequence into account. Consequently, as the expressions become longer, the differences between the embeddings of neighbouring tokens also decrease, as their meanings become increasingly similar relative to the expanding overall context.

## Corpus analysis

As the qualitative results from the visualisations suggested multiple possibly under-utilised layers in GPT-2, we conduct the corpus analysis and utilisation-based pruning with GPT-2. The corpus-level analysis on the Brown validation split reveals a non-uniform utilisation profile across the 24 transformer layers of GPT-2, which is shown in Table 1. The token-to-token utilisation is low at the bottom and top of the stack with less than 10%. The utilisation rates increase towards the central layers and reach the maximum of 55.42% at layer 13. As the layers' utilisation is determined for a corpus, this pattern is consistent for various texts and does not depend on specific linguistic or sematic features.

| | | | |
|---|---|---|---|
| $U_1 = 12.40\%$ | $U_2 = 6.45\%$ | $U_3 = 3.26\%$ | $U_4 = 2.24\%$ |
| $U_5 = 2.10\%$ | $U_6 = 2.94\%$ | $U_7 = 11.29\%$ | $U_8 = 38.85\%$ |
| $U_9 = 42.07\%$ | $U_{10} = 42.89\%$ | $U_{11} = 48.57\%$ | $U_{12} = 54.72\%$ |
| $U_{13} = 55.42\%$ | $U_{14} = 51.86\%$ | $U_{15} = 47.80\%$ | $U_{16} = 44.86\%$ |
| $U_{17} = 42.87\%$ | $U_{18} = 40.39\%$ | $U_{19} = 28.42\%$ | $U_{20} = 8.75\%$ |
| $U_{21} = 3.06\%$ | $U_{22} = 2.03\%$ | $U_{23} = 1.75\%$ | $U_{24} = 1.57\%$ |

**Table 1. Utilisation rates of the 24 layers of GPT-2 for the Brown corpus.** Utilisation rate for each layer of GPT-2 (Radford et al., 2019) for the scoring split of the Brown corpus. The utilisation rate shows the portion of information flow that is directed in the token-to-token direction. Low utilisation rates indicate little change between tokens.

The observed layer-wise profile for a corpus suggests that GPT-2 allocates information unevenly across its layers. GPT-2 appears to concentrate its most active behaviour in a mid-depth region while leaving considerable portions of the lower and upper stack less engaged. Thus, the corpus analysis with DONALD-D confirms the qualitative results of the visualisation as it does not point to single isolated layers but hints at a broader redundancy structure across multiple layers.

## Model pruning

We apply the utilisation rates determined by DONALD-D for pruning GPT-2. As the threshold for a layer's minimum utilisation rate can be freely chosen, we conduct the pruning for various thresholds to assess the threshold's influence on parameter compression, speedup, and performance loss. The results of our pruning experiments for GPT-2 are shown in Table 2.

| Utilisation threshold for pruning | Number of remaining layers | Parameter compression | Speedup | Performance loss |
|---|---|---|---|---|
| 20% | 12 | 42.60% | 73.59% | 19.14% |
| 25% | 12 | 42.60% | 72.57% | 19.14% |
| 30% | 11 | 46.15% | 107.05% | 18.88% |
| 35% | 11 | 46.15% | 104.77% | 18.88% |
| 40% | 10 | 49.70% | 123.80% | 20.22% |
| 45% | 5 | 67.45% | 261.90% | 19.51% |
| 50% | 3 | 74.55% | 534.69% | 18.73% |

**Table 2. Model complexity, throughput and performance loss when pruning GPT-2 at various utilisation thresholds.** Metrics for versions of GPT-2 (Radford et al., 2019) that were pruned according to the utilisation rates determined by DONALD-D. Each pruned model was readjusted with the full model as a teacher. Each row shows the results for one pruned model and the first column shows the applied utilisation threshold when pruning the model. The other columns show how many of the original model's 24 layers remain after pruning, and what portion of the original model's parameters is pruned. The speedup indicates how many more tokens per second the model can process after pruning, while the performance loss shows the pruning-caused decrease in average negative log-likelihood for the test split.

The results show that structural pruning based on DONALD-D's utilisation scores produces considerable reductions in model size. Pruning the 24 layers of GPT-2 with various thresholds results numbers of retained layers ranging from 12 down to 3, corresponding to a transformer depth reduction of 50% to 87.5%. However, as the embedding layers, positional embeddings, and output head are preserved during the utilisation-based pruning, the depth reduction does not directly correspond to actual parameter compression. Pruning at the various thresholds leads to total parameter compression between 42.60% and 74.55% compared to the unpruned model.

The parameter compression entails increased throughput as less computations must be performed. The inference throughput, measured in processed tokens per second, increases by 73.59% for the least restrictive utilisation threshold and by 534.69% for the most restrictive pruning. These results indicate that the removed layers account for a considerable portion of computational cost. The substantial increase in inference also shows that DONALD-D-based pruning can deliver strong efficiency gains while keeping the model structurally faithful to the original architecture.

Despite the considerable reduction in model complexity and computational cost, the performance loss on the held-out Brown test split remains in a narrow band. Across all reported thresholds, the pruning-induced decrease in average negative log-likelihood is between 18.73% and 20.22%. Thus, retaining the most-utilised 12 layers or just the 3 most-utilised layers does not result in a proportional collapse in performance. Instead, the pruned students converge to a stable performance after the distillation-based readjustment procedure.

This result, within the limits of our distillation procedure and corpus, suggests that DONALD-D exposes plateaus of redundant transformer blocks and not merely threshold-specific compression points. The pruned model's performance depends more on the structural compression as a whole than on the precise number of retained layers. Thus, the DONALD-D-pruned model retains baseline performance even after substantial pruning. This allows for reductions in model complexity and computational cost that are disproportionately greater than the degradation in language modelling quality.

# Discussion

In this work, we use diffusion tensor imaging to visualise information flow between tokens in natural language expressions. The differences in information flow patterns between LLMs show how representations of natural language expressions differ from model to model. Even for models with comparable architectures, like the encoder-only models BERT and Longformer, DONALD-D reveals differences in the embedding space. In the case of BERT, for which the function of different layer sections is established (Jawahar et al., 2019), DONALD-D shows how the highest four layers capture surface features like punctuation, as only for these tokens there is a token-to-token information flow.

The results of our corpus analysis show that DONALD-D can be used to signal opportunities for pruning as it identifies a considerable amount of depth redundancy in GPT-2-medium. The method produces a traceable, deterministic, and mathematically well-defined utilisation ranking across layers. Using the utilisation ranking for pruning the most redundant layers of an LLM provides considerable reductions in parameter count and strong inference speedups while keeping the performance loss disproportionately small. Thus, the empirical finding is that DONALD-D-based pruning comes at a cost, but that a large fraction of the model's depth can be removed with only a moderate degradation in performance.

A limitation of DONALD-D is that it collapses the hidden units of each layer and only considers the arithmetic mean. This simplification allows for clear visualisations that are interpretable.

However, collapsing the hidden units into one value presents a simplification that potentially obscures changes in the hidden units as only their mean is considered. Especially for pruning, considering all the hidden layers separately and creating three-dimensional diffusion models could provide more detailed information on the utilisation rates of an LLM's layers.

DONALD-D provides an extension of existing point-plot visualisations. Established visualisations of word embeddings do not consider a word together with its surrounding context or various neighbouring words simultaneously. The average meaning extracted by an LLM is plotted together with other words' embeddings and their relative positions might reveal systematic differences in how an LLM represents meaning (Mikolov et al., 2013). With DONALD-D, the information flow between word embeddings can be visualised for actual natural language expressions. This visualisation provides insights into how an LLM's word embeddings function within the context of natural language. DONALD-D can be used to analyse natural language expressions of any length. However, as DONALD-D highlights the differences in neighbouring word embeddings, utilisation rates tend to decrease as analysed expressions get longer. We leave it to future research to show whether visualisations from DONALD-D can assist in other scientific fields, e.g., with linguistic analyses or the improvement of LLMs for certain NLP tasks. Beyond the visualisations, the underlying information flows remain clearly interpretable, enabling quantitative analyses on entire corpora. Thus, DONALD-D presents a novel visualisation and analysis method that enhances the interpretability of LLMs and reveals opportunities for pruning. We hope that our work will open up a new avenue for the analysis and understanding of LLMs.

**Acknowledgements**

None.